\documentclass[11pt,a4paper]{article}
\usepackage[hyperref]{acl2019}
\usepackage{times}
\usepackage{latexsym}
\usepackage{import}
\usepackage{graphicx}
\usepackage[caption=false]{subfig}
\usepackage{amsmath}
\usepackage{amssymb}
\usepackage{bm}

\usepackage{url}

\aclfinalcopy 

\title{Correlating Twitter Language with Community-Level Health Outcomes}

\author{Arno Schneuwly \\
  EPFL \\
  \texttt{arno.schneuwly@epfl.ch} 
  \And
  Ralf Grubenmann \\
  SpinningBytes \\
  \texttt{rg@spinningbytes.com} 
  \AND 
  \bf{S\'everine Rion Logean}  \\
  Swiss Re \\
  \texttt{severine\_rion@swissre.com}\hspace{-8mm}
  \And
  \bf{Mark Cieliebak}  \\
  ZHAW \\
  \texttt{ciel@zhaw.ch}\hspace{-8mm}
  \And
  \bf{Martin Jaggi} \\
  EPFL \\
  \texttt{martin.jaggi@epfl.ch}\hspace{-2mm}
  }

\date{}

\begin{document}
\maketitle
\begin{abstract}
We study how language on social media is linked to diseases such as atherosclerotic heart disease (AHD), diabetes and various types of cancer.
Our proposed model leverages state-of-the-art sentence embeddings, followed by a regression model and clustering, without the need of additional labelled data. It allows to predict community-level medical outcomes from language, and thereby potentially translate these to the individual level. The method is applicable to a wide range of target variables and allows us to discover known and potentially novel correlations of medical outcomes with life-style aspects and other socioeconomic risk factors.
\end{abstract}

\section{Introduction}
Surveys and empirical studies have long been a cornerstone of psychological, sociological and medical research, but each of these traditional methods pose challenges for researchers. They are time-consuming, costly, may introduce a bias or suffer from bad experiment design.

With the advent of big data and the increasing popularity of the internet and social media, larger amounts of data are now available to researchers than ever before. This offers strong promise new avenues of research using analytic procedures, obtaining a more fine-grained and at the same time broader picture of communities and populations as a whole~\cite{salathe2018digital}. Such methods allow for faster and more automated investigation of demographic variables. It has been shown that Twitter data can predict atherosclerotic heart-disease risk at the community level more accurately than traditional demographic data~\cite{eichstaedt2015psychological}. The same method has also been used to capture and accurately predict patterns of excessive alcohol consumption~\cite{curtis2018can}.

In this study, we utilize Twitter data to predict various health target variables (AHD, diabetes, various types of cancers) to see how well language patterns on social media reflect the geographic variations of those targets. Furthermore, we propose a new method to study social media content by characterizing disease-related correlations of language, by leveraging available demographic and disease information on the community level. In contrast to~\cite{eichstaedt2015psychological}, our method is not relying on word-based topic models, but instead leverages modern state-of-the-art text representation methods, in particular sentence embeddings, which have been in increasing use in the Natural Language Processing, Information Retrieval and Text Analytics fields in the past years. We demonstrate that our approach helps capturing the semantic meaning of tweets as opposed to features merely based on word frequencies, which come with robustness problems \cite{brown2018does,schwartz2018more}. We examine the effectiveness of sentence embeddings in modeling language correlates of the medical target variables (disease outcome).

Section~\ref{sec:method} gives a generalized description of our method. We apply the previously described method to the tweets and health data in Section~\ref{sec:datasources} The system's performance is evaluated in Section~\ref{sec:results} followed by the discussion in Section~\ref{sec:discussion}. Our code is available on \href{https://github.com/epfml/correlating-tweets}{github.com/epfml/correlating-tweets}.

\section{Method}\label{sec:method}
We are given a large quantity of text (sentences or tweets) in the form of social media messages by individuals. Each individual---and therefore each sentence---is assigned to a predefined category, for example a geographic region or a population subset. We assume the number of sentences to be significantly larger than the number of communities. 
Furthermore, we assume that the target variable of interest, for example disease mortality or prevalence rate, is available for each community (but not for each individual). Our system consists of two subsystems:

\begin{enumerate}
    \item \emph{(Prediction)}
    The predictive subsystem makes predictions of target variables (e.g. AHD mortality rate) based on aggregated language features. The resulting linear predictions are applicable on the community level (e.g. counties) or on the individual level, and are trained using k-fold cross-validated Ridge regression.
    \item \emph{(Interpretability)}
    The averaged regression weights from the prediction system allow for interpretation of the system:
    We use a fixed clustering (which was obtained from all sentences without any target information), and then rank each topic cluster with respect to a prediction weight vector from point 1).
    The top and bottom ranked topic clusters for each target variable give insights into known and potentially novel correlations of topics with the target medical outcome.
\end{enumerate}

In summary, the community association is used as a proxy or weak labelling to correlate individual language with community-level target variables.
The following subsections give a more detailed description of the two subsystems.

\subsection{System Description}
Let $\mathcal{S}$ be the set of sentences (e.g. tweets), with their total number denoted as $|\mathcal{S}| = S$. Each sentence is associated to exactly one of the $A$ communities $\mathcal{A}=\{a_1,\dots,a_A\}$ (e.g. geographic regions). The function $\delta: \mathcal{S} \rightarrow \mathcal{A}$ defines this mapping. Let $\mathbf{y} \in \mathbb{R}^{A}$ be the target vector for an arbitrary target variable, so that each community $a_j$ has a corresponding target value $y_{a_j}\in\mathbb{R}$. 

\begin{figure}
\includegraphics[width=7.5cm]{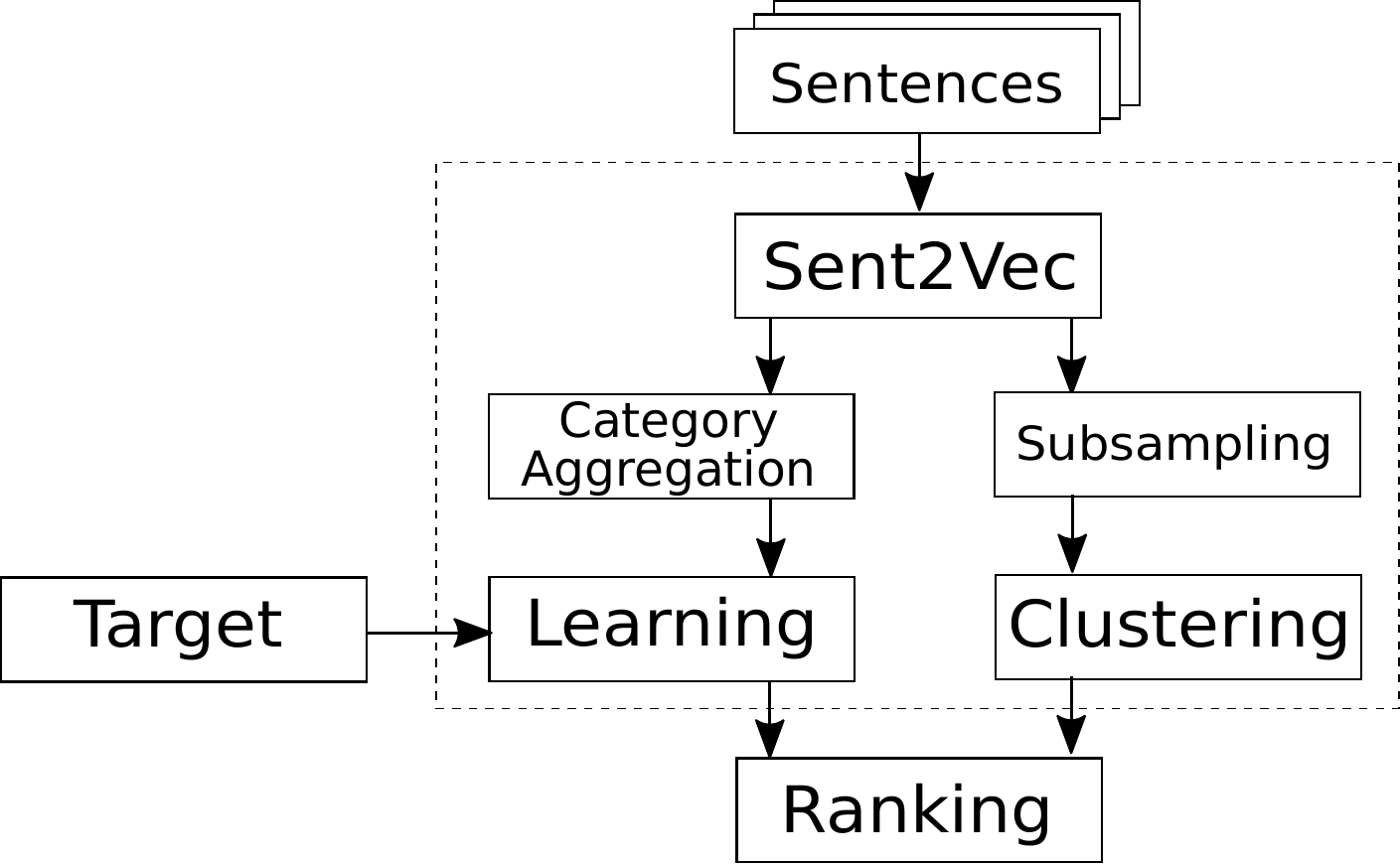}
\caption{System Description.}
\label{fig:systemDescription}

\end{figure}

\paragraph{Preprocessing and Embeddings.}
The complete linguistic preprocessing pipeline of a sentence is incorporated by the function $\rho(s_i), \ \forall \ i \in \{1,\dots,S\}$, which represents an arbitrary sentence~$s_i$ as a sequence of tokens. Each sentence~$s_i$ then is represented by a $D$-dimensional embedding vector providing a numerical representation of the semantics for the given short text:
\begin{equation}
\mathbf{x}_i = \textit{Sent2Vec}(\rho(s_i)) \in \mathbb{R}^{D}.
\end{equation}
While our method is generic for any text representation method, here Sent2Vec \cite{pgj2017unsup} was chosen for its computational efficiency and scalability to large datasets.

\subsection{Feature Aggregation}
We use averaging of the sentence embedding vectors over each community to obtain the language features for each community.
Formally, the complete feature matrix of all sentences is denoted as~$\textbf{X} \in \mathbb{R}^{S\times D}$.
For our approach, the sentence embedding features 
are averaged over each community $a_j$.
Formally, an individual feature $\overline{x}_{{a_j},d}$ of the averaged embedding $\overline{\mathbf{x}}_{a_j} \in \mathbb{R}^{1 \times D}$ for a given community~$a_j$ is defined as
\begin{equation}
\overline{x}_{{a_j},d} = \frac{1}{N_{a_j}} \sum_{x_i : s_i \in \mathcal{S} \ \ \land \ \  \delta(s_i) = a_j} x_{i,d},
\end{equation}
where $N_{a_j} = |\{s_i: s_i \in S \land \delta(s_i) = a_j\}|$ is the number of sentences belonging to community $a_j$. Consequently, the aggregated community-level embedding matrix is given by
\begin{equation}
\overline{\mathbf{X}} = \begin{bmatrix}
\overline{\mathbf{x}}_{a_1}^{\top} \\
\vdots \\
\overline{\mathbf{x}}_{a_A}^{\top}
\end{bmatrix} \in \mathbb{R}^{A \times D}.
\end{equation}

\subsection{Train-Test Split}
Leveraging the targets available for each community, our regression method is applied to the aggregated features $\overline{\mathbf{X}}$ and the target $\mathbf{y}$. We employ $K$-fold cross-validation: the previously defined set $\mathcal{A}$ is split into K as equally sized pairwise disjoint subsets $\mathcal{A}_k$ as possible such that: $\mathcal{A}=\bigcup_{k=1}^K \mathcal{A}_k$, $\mathcal{A}_i \cap \mathcal{A}_j = \emptyset \ \forall i,j \in {1,\dots, K}, i \neq j$ and $|\mathcal{A}_1| \approx \dots \approx|\mathcal{A}_K|$. The training set for a fold~$k$ is $\text{TR}_k = \left(\bigcup_{i=1}^K \mathcal{A}_i \right)\setminus \mathcal{A}_k$ 
with the corresponding test set $\text{TE}_k = \mathcal{A}_k$, where $N^{\theta}_k = |\text{TR}_k|$ and $N^{\Lambda}_k=|\text{TE}_k|$. The operators $\theta_k: \{1,\dots,N^{\theta}_k\} \rightarrow \text{TR}_k$ and $\Lambda_k: \{1,\dots,N^{\Lambda}_k\} \rightarrow \text{TE}_k$ uniquely map the indexes to the corresponding communities $a_j$ for the $k^{th}$ train-test split. For each split $k$ the train and test embedding matrices respectively are defined as 
\begin{equation}
  \overline{\mathbf{X}}_{\theta_k} = \left[\overline{\mathbf{x}}_{\theta_k(1)},\dots,\mathbf{\overline{x}}_{\theta_k(N_k^{\theta})} \right]^{\top}, \vspace{-2mm}
\end{equation}
\begin{equation}
  \overline{\mathbf{X}}_{\Lambda_k} = \left[\overline{\mathbf{x}}_{\Lambda_k(1)},\dots,\mathbf{\overline{x}}_{\Lambda_k(N_k^{\Lambda})} \right]^{\top}.
\end{equation}

Accordingly, we define the target vectors 
\begin{equation}
    \mathbf{y}_{\theta_k} = \left[y_{\theta_k(1)}, \dots, y_{\theta_k(N_k^{\theta})} \right]^{\top},\vspace{-3mm}
\end{equation}
\begin{equation}
\mathbf{y}_{\Lambda_k} = \left[ y_{\Lambda_k(1)}, \dots, y_{\Lambda_k(N_k^{\Lambda})} \right]^{\top}.   
\end{equation}

\subsection{Ridge Regression}
For each train-test split $k$ we perform linear regression from the community-level textual features $\overline{\mathbf{X}}_{\theta_k}$ to the health target variable $\mathbf{y}_{\theta_k}$. We employ Ridge regression \cite{hoerl1970ridge}.
In our context, the Ridge regression is defined as the following optimization problem:
\begin{equation}
\underset{\bm{\omega}_k\in \mathbb{R}^D}{\min} \  \frac{1}{2A}\sum_{i=1}^{N^\theta_k} \big[y_{\theta_k(i)} - \mathbf{\overline{x}}_{\theta_k}^\top\bm{\omega}_k\big]^2 + \lambda\|\bm{\omega}_k\|_2^2, 
\end{equation}
where the optimal solution is
\begin{equation}
    \bm{\omega}_k^\star = \big(\bm{\overline{X}}_{\theta_k}^{\top}\mathbf{\overline{X}}_{\theta_k} + 2N_{k}^\theta\lambda\mathbf{I}\big)^{-1}\mathbf{\overline{X}}_{\theta_k}^{\top} \quad \in \mathbb{R}^D.
\end{equation}

Within each each fold we tune the regularization parameter $\lambda$.

\subsection{Prediction Subsystem}\label{sec:predsystem}
Let $\bm{\overline{y}}_{\Lambda_k} = \mathbf{\overline{X}}_{\Lambda_k} \bm{\omega}_k^\star = [\overline{y}_{\Lambda_k(1)}, \dots, \overline{y}_{\Lambda_k(N^{\Lambda}_k)}]^{\top}$ be the predicted values for the test set of the split~$k$. The concatenated prediction vector for all splits is
\begin{equation}
\mathbf{\overline{y}}_\Lambda = \begin{bmatrix}
\overline{\mathbf{y}}_{\Lambda_1}^{\top} \\
\vdots \\
\overline{\mathbf{y}}_{\Lambda_K}^{\top}
\end{bmatrix}  \in \mathbb{R}^A 
\end{equation}

Accordingly, we define the concatenated true target vector as
\begin{equation}
\mathbf{y}_\Lambda = \begin{bmatrix}
\mathbf{y}_{\Lambda_1}^{\top} \\
\vdots \\
\mathbf{y}_{\Lambda_K}^{\top}
\end{bmatrix} \in \mathbb{R}^A,
\end{equation}
i.e., the set of individual scalars is identical to the entries in the original target vector~$\bm{y}$. 
The predictive performance of the system can be assessed through the following metrics:
\begin{itemize}
    \item Pearson Correlation Coefficient
    \item Mean Average Error of prediction (MAE)
    \item Classification Accuracy for Quantile Prediction
\end{itemize}

The first two metrics are evaluated with the vectors $\overline{\mathbf{y}}_\Lambda$ and $\mathbf{y}_\Lambda$ from all folds. In the quantile-based assessment we independently bin the true values $\mathbf{y}_\Lambda$ and the predicted values $\overline{\mathbf{y}}_\Lambda$ into $C$ different quantiles. Each individual true and predicted value is assigned to a quantile $c_j \in \{c_1, \dots, c_C\}$. These assignments can be used to visually compare results on a heat-map or as regular evaluation scores in terms of accuracy.

\subsubsection{Ridge-Weight Aggregation}
For the final prediction model, the regression weights $\bm{\omega}^\star_k$ from Ridge regression are averaged over the $K$  folds, i.e. $\bm{\overline{\omega}} = \frac{1}{K} \sum_{k=1}^K \bm{\omega}^\star_k$. 

For every sentence embedding $\mathbf{x}_q$, the prediction is computed as $\overline{y}_q = \mathbf{x}_q^{\top} \bm{\overline{\omega}} \in \mathbb{R}$.

\subsection{Interpretation Subsystem: Cluster Ranking}
We employ predefined textual topic clusters---which are independent of any target values---in order to enable interpretation of the textual correlates.
Each cluster is a collection of sentences and should, intuitively, be interpretable as a topic, e.g. separate topics about indoor and outdoor activities as shown in Fig.~\ref{fig:melanoma}.
For each cluster $m$ a ranking score can be computed with respect to a linear prediction model $\bm{\overline{\omega}}$ such as defined above. Let $\mathcal{Q}_m = \{ q: \zeta(q) = m \ \land \ q \  \in \mathcal{Q}\}$ be the set of sentences assigned to cluster $m$. The score $\iota_m$ for the cluster $m$ is the average of all predictions $\overline{y}_q = \mathbf{x}_q^{\top} \bm{\overline{\omega}}$
within the cluster $m$:
\begin{equation}
\iota_m = \frac{1}{|\mathcal{Q}_m|} \sum_{\overline{y}_q:\ q \in \mathcal{Q}_m} \overline{y}_q
\end{equation}

By ordering the scores $\iota_m$ of all clusters, we obtain the final ranking sequence of all clusters, with respect to the target-specific model $\bm{\overline{\omega}}$.
\\

\emph{Clustering Preprocessing.}
For obtaining the fixed clustering, as $\mathbf{X}$ is a very large matrix, clustering might require subsampling to reduce computational complexity. Hence, $Q$ out of the $S$ embeddings in $\mathcal{S}$ are randomly subsampled into the set $\mathcal{Q}$. The mapping $\Phi(Q) = [\phi(1), \dots, \phi(Q)]^{\top}$ is a uniformly random selection of row indexes in $\mathbf{X}$ out of $\binom{N}{Q}$. We define the subsampled data matrix as $\mathbf{X}_Q = \left [ \mathbf{x}_{\phi (1)}, \dots, \mathbf{x}_{\phi (Q)} \right]^{\top} \in \mathbb{R}^{Q \times D}$.

The subset $\mathbf{X}_Q$ is clustered with the Yinyang K-Means algorithm \cite{kmeans}. We use~$M$ centroids and the cosine similarity as a distance function. The cluster assignment vector $\mathbf{M} \in [1,\dots,M]^{\top}$ assigns one cluster for each embedding in $\mathbf{X}_Q$. Accordingly, the operator $\zeta:~\{1,\dots,Q\}~\rightarrow~\{1,\dots,M\}$ indicates the assigned cluster $m$ for a given sentence $s$ in $\mathcal{Q}$ (see cluster ranking above). The cluster centers are defined in $\mathbf{M}_Q \in \mathbb{R}^{Q \times D}$.

\section{Data sources}\label{sec:datasources}
We apply the method described in Section~\ref{sec:method} to the following setting: The pool of sentences $\mathcal{S}$ consists of geotagged Tweets. The assigned locations are in the United States. The geotags are categorized into US-counties which represent the set of communities $\mathcal{A}$. The target variables~$\mathbf{y}$ are health-related variables, for example normalized mortality or prevalence rates. We focus on cancer and AHD mortality as well as on diabetes prevalence. Hence, the quantile-based predictions give a categorization of the Ridge regression predictions on a US-county level. The ranked topics assess what language might relate to higher or lower rates of the corresponding disease. Table~\ref{tab:datasources} provides an overview of the size of the data sources, the year the data was collected in and the mean $\mu$ and standard deviation $\sigma$ of the target variables. Not all counties are covered in the publicly available datasets, usually being limited to more populous counties.
The collected Tweets are from 2014 and 2015. The target variables are the union-averaged values from 2014 and 2015: if the target variable is available for both years the two values are averaged. Conversely, if a county data point is only available for one, but not both years, we use this standalone value. 

\begin{table}[h]
\setlength{\tabcolsep}{5pt}
\vspace{2mm}
\centering
\begin{tabular}{ |l|r|r|c| } 
\hline
\textbf{Name} & \textbf{\# tweets} & \textbf{Year} & \\ 
\hline
\hline
Datorium & 147M & 14/15 & \\
\hline
\hline
\textbf{Name} & \textbf{\# counties} & \textbf{Year} &  $\bm{\mu}$, $\bm{\sigma}$\\ 
\hline
\hline
AHD & 803 & 14/15 & 43.0, 16.1 \\
\hline
Diabetes & 3129 & 13 & 9.7, 2.2 \\
\hline
Breast & 487 & 13/14 & 12.4, 2.8 \\
\hline
Colon &  490 & 13/14 & 12.1, 3.0 \\
\hline
Liver & 293 & 13/14 & 7.5, 2.4 \\
\hline
Lung &  1612 & 13/14 & 52.4, 16.2 \\
\hline
Melanoma & 162 & 13/14 & 3.8, 1.2 \\
\hline
Prostate &  351 & 13/14 & 8.5, 2.0 \\
\hline
Stomach &  136 & 13/14 & 3.6, 0.9 \\
\hline
\end{tabular}
\caption{Overview of data sources. }
\label{tab:datasources}
\end{table}

\subsection{Datorium Tweets}
Tweets are short messages of no more than 140 characters\footnote{Twitter increased the limit to 280 characters in 2017, which doesn't affect our data.} published by users of the Twitter platform. They reflect discussions, thoughts and activities of its users. We use a dataset of approximately 144 million tweets collected from first of June 2014 to first of June 2015 \cite{datorium}. Each tweet was geotagged by the submitting user with exact GPS coordinates and all tweets are from within the US, allowing accurate county-level mapping of individual tweets.

\subsection{AHD \& Cancer Mortality}
Our source of the statistical county-level target variables is the CDC WONDER\footnote{US Centers for Disease Control and Prevention - Wide-ranging Online Data for Epidemiologic Research.} database \cite{wonder} for AHD and cancer. Values are given as deaths per capita (100'000).

\subsection{Diabetes Prevalence}
We use county-wise age-adjusted diabetes prevalence data from the year 2013 \cite{diabetes}, provided as percent of the population afflicted with type II diabetes. The data is available for almost all the 3144 US counties, making it a valuable target to use.

\section{Results}\label{sec:results}

The results of our method for the various target variables are listed in Table~\ref{tab:results} along with the performance of the baseline model outlined in Section~\ref{sec:baseline}. We provide the Pearson correlation ($\bm{\rho}$) and the mean absolute error (MAE) of our system along with the baseline model's Pearson correlation. 

\subsection{LDA Baseline Model}\label{sec:baseline}
We reimplemented the approach proposed by \citet{eichstaedt2015psychological} as a baseline for comparison, and were able to reproduce their findings about AHD with recent data: similar results were found with the Datorium Twitter dataset \cite{datorium} and CDC AHD data from 2014 and 2015. Their approach averages topics generated with Latent Dirichlet Allocation (LDA) of tweets per county as features for Ridge regression. We do not use any hand-curated emotion-specific dictionaries, as these did not impact performance in our experiments. We used the predefined \textit{Facebook} LDA coefficients of \citet{eichstaedt2015psychological}, updated them with the word frequencies of our collected Twitter data \cite{datorium}. Our results are computed with a 10-fold cross-validation and without any feature selection.

\begin{table}[h]
\centering
\begin{tabular}{ |l|c|c|c| } 
\hline
\textit{Type} & $\bm{\rho}$ & $\bm{\rho}$ \textit{LDA} & \textit{MAE}  \\ 
\hline
\hline
AHD & \textbf{0.46} & 0.31 & 13.4 \\
\hline
\hline
Diabetes & \textbf{0.73} & 0.72 & 1.1 \\
\hline
\hline
Breast & \textbf{0.44} & 0.42 & 1.80  \\
\hline
Colon & \textbf{0.55} & 0.51 & 1.87  \\
\hline
Liver & 0.29 & \textbf{0.40} & 1.59  \\
\hline
Lung & \textbf{0.68} & 0.63 & 8.44  \\
\hline
Melanoma & \textbf{0.72} & 0.61 & 0.68  \\
\hline
Prostate & \textbf{0.39} & 0.38 & 1.34  \\
\hline
Stomach & 0.44 & \textbf{0.51} & 0.72  \\
\hline
\end{tabular}
\caption{Results of predictions on different health targets. $\bm{\rho}$: our system (Section~\ref{sec:predsystem}), $\bm{\rho}$ LDA: topic model baseline (\citet{eichstaedt2015psychological}, Section~\ref{sec:baseline}), MAE: mean absolute error of our system (Section~\ref{sec:predsystem}).}
\label{tab:results}
\end{table}

\begin{figure*}[!htb]
    \centering
    \vspace{2mm}
    \subfloat[]{
        \includegraphics[width=0.45\linewidth]{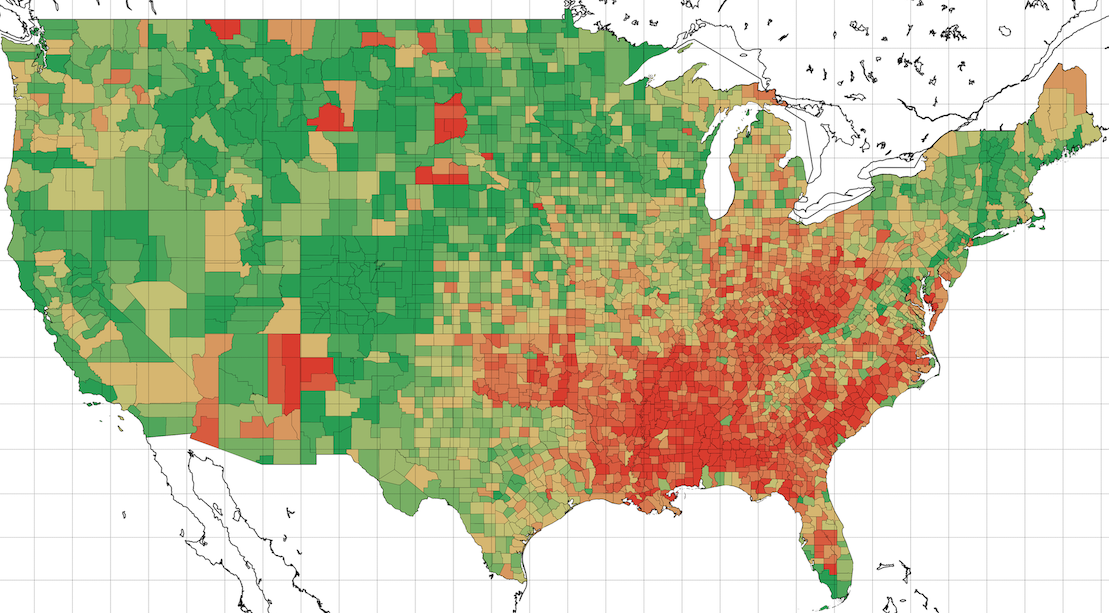}
    }
    ~
    \subfloat[]{
        \includegraphics[width=0.45\linewidth]{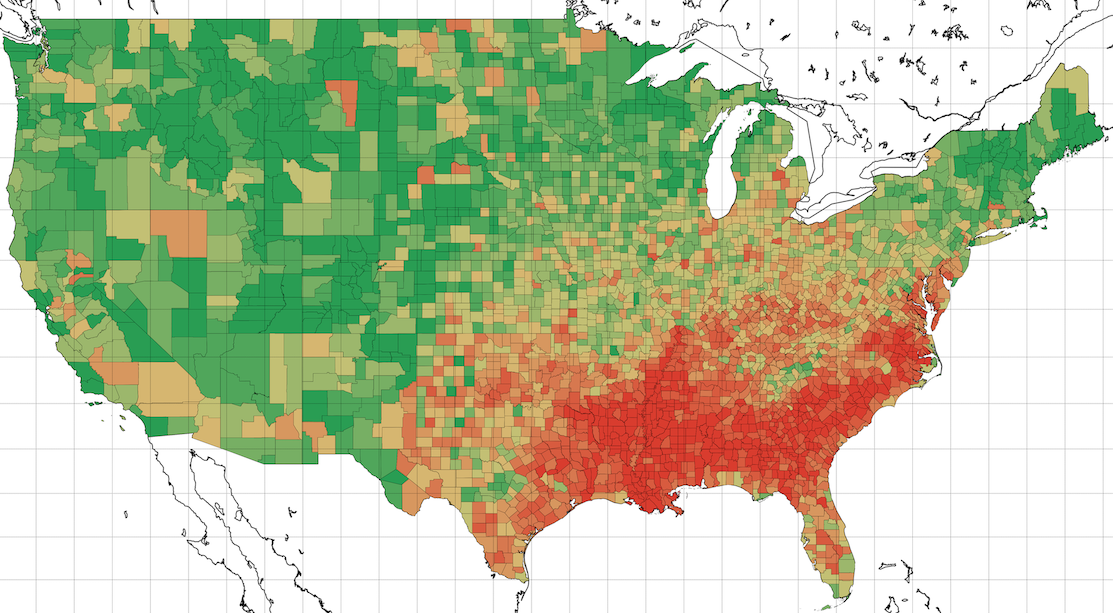}\vspace{-2mm}
    }
  \caption[Diabetes]{Quantiles of the prevalence of \textbf{diabetes}. (a) Target values (b) Predicted values from tweets}
  \label{fig:diabMap}
\end{figure*}

\begin{figure}[!htb]

\subfloat[]{
   \includegraphics[width=0.45\linewidth]{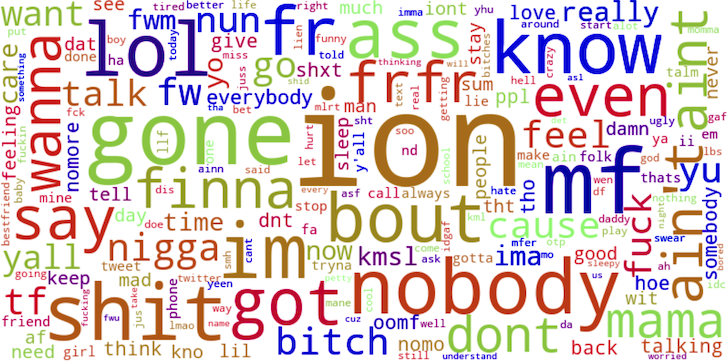}
}
\subfloat[]{
   \includegraphics[width=0.45\linewidth]{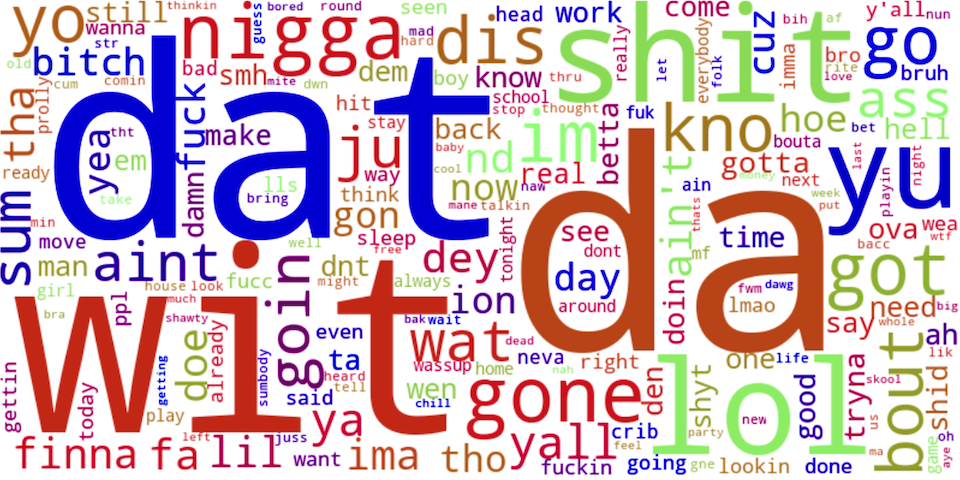}
}

\subfloat[]{
   \includegraphics[width=0.45\linewidth]{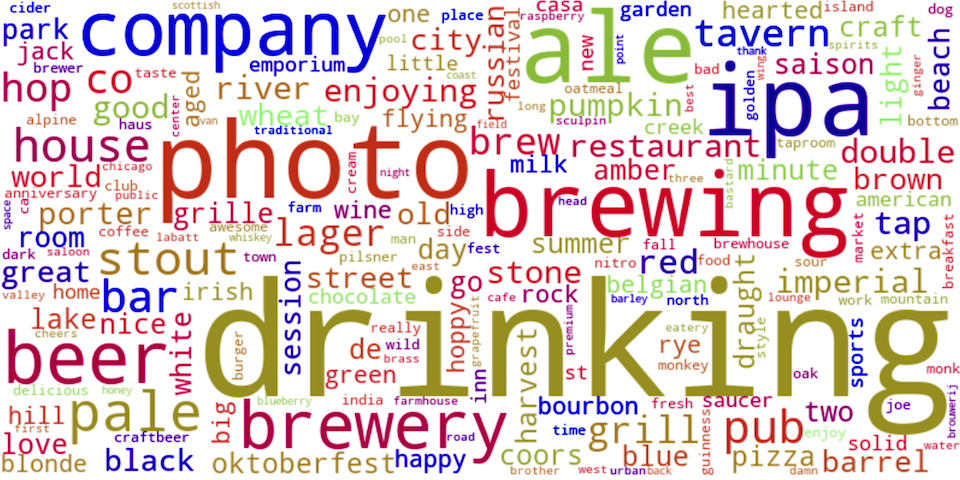}
}
\subfloat[]{
   \includegraphics[width=0.45\linewidth]{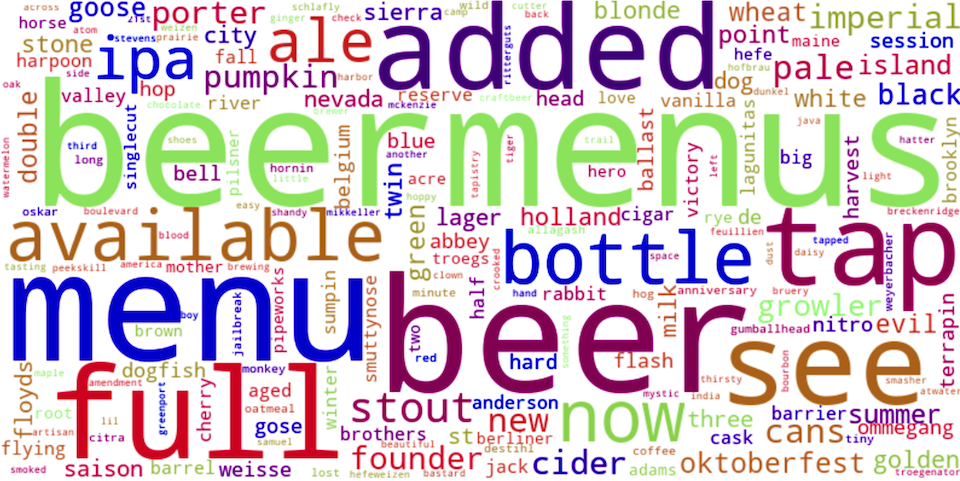}
}

\caption[Diabetes]{Word clouds of topics correlating with \textbf{diabetes}: (a) (b) strongest positive correlation (c) (d) strongest negative correlation among $M=2000$ clusters.\vspace{-2mm}}
\label{fig:diabeteswords}
\end{figure}

\begin{figure}[!htb]

\subfloat[]{
   \includegraphics[width=0.45\linewidth]{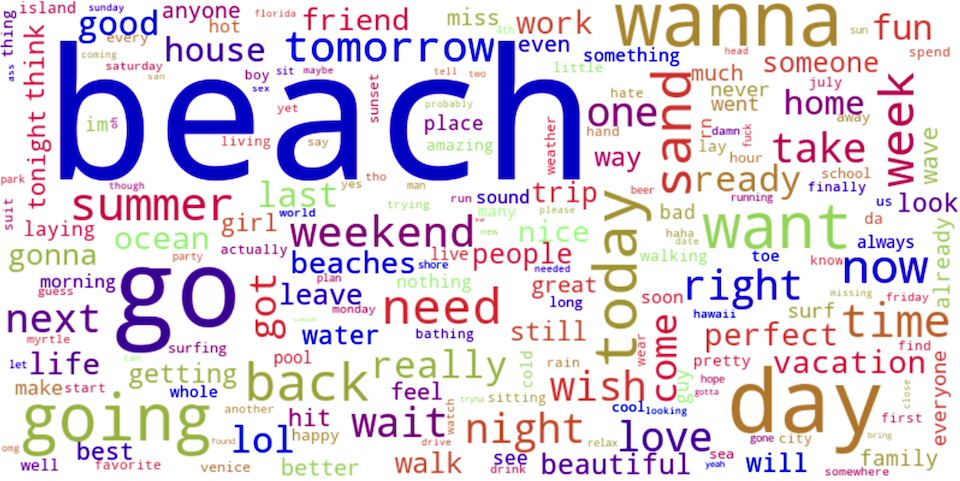}
}
\subfloat[]{
   \includegraphics[width=0.45\linewidth]{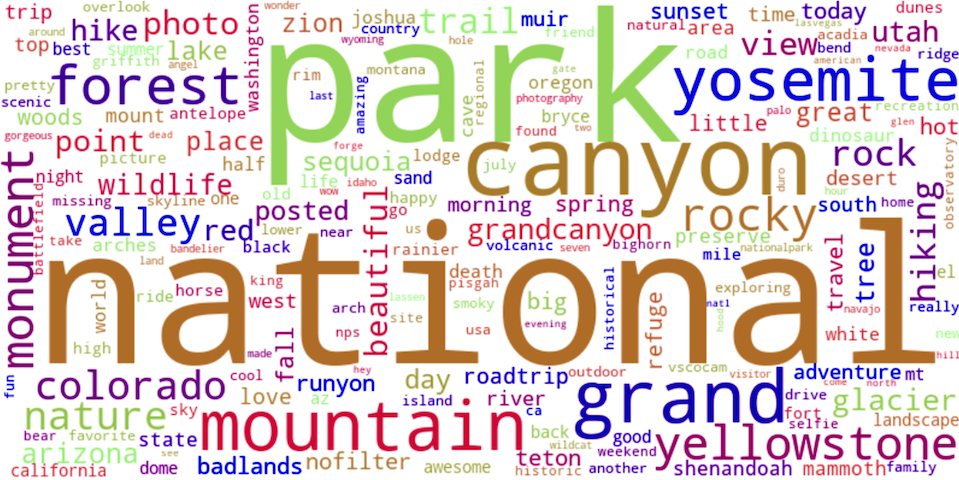}
}

\subfloat[]{
   \includegraphics[width=0.45\linewidth]{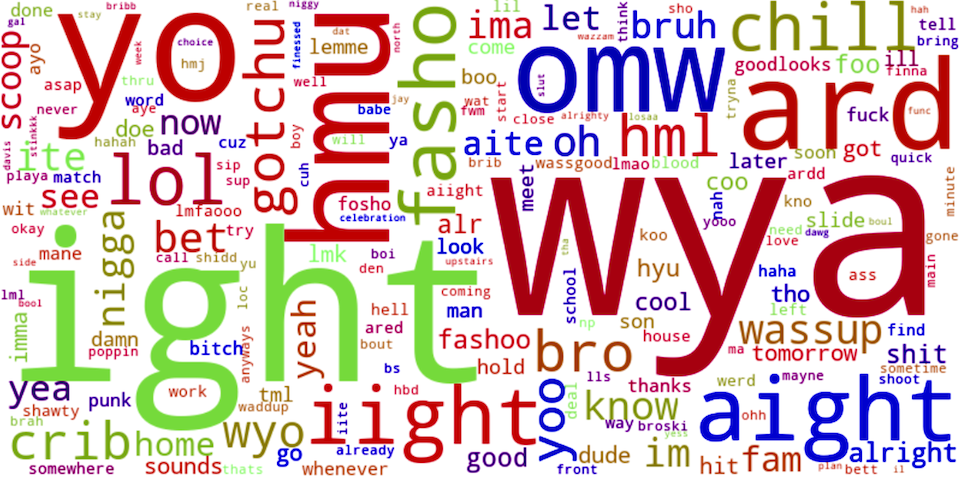}
}
\subfloat[]{
   \includegraphics[width=0.45\linewidth]{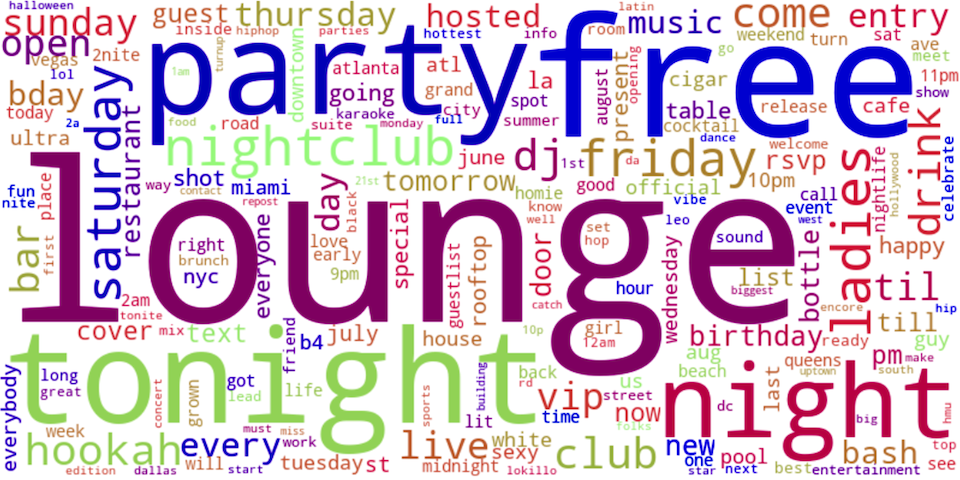}
}

\caption[Melanoma]{Word clouds of topics correlating with \textbf{melanoma}: (a) (b) strongest positive correlation (c) (d) strongest negative correlation among $M=2000$ clusters.}
\label{fig:melanoma}
\end{figure}

\subsection{Detailed Results}
In this section we discuss a selection of our results in detail, with additional information available in~Appendix~\ref{sec:addCharts}.

Diabetes has a strong demographic bias, with a higher prevalence in the south-east of the US, the so called \textit{diabetes belt}. Compared to the national average, the african-american population in the diabetes belt has a higher risk of diabetes by a factor of more than 2 \cite{barker2011geographic} and the south-east of the US has a large african-american population. Therefore, linguistic features~\cite{green2002african} common in african-american are a strong predictor of diabetes rates. The model learns these linguistic features, as seen in Figure~\ref{fig:diabeteswords}, and its predictions closely match the actual geographic distribution, as seen in Figure~\ref{fig:diabMap}. A moderate alcohol consumption is linked to a low risk of type II diabetes compared to no or excessive consumption \cite{koppes2005moderate}. The strongest negatively correlated word clouds in Figure \ref{fig:diabeteswords} support this finding.

The most positively related word clouds for melanoma in Figure~\ref{fig:melanoma} are related to outdoor activities \cite{elwood1985cutaneous}. Conversely, the strongest negatively correlated word clouds suggest indoor activity related language.

\section{Discussion}\label{sec:discussion}
In this paper, we introduced a novel approach for language-based predictions and correlation of community-level health variables. For various health-related demographic variables, our approach outperforms in most cases (Table~\ref{tab:results}) similar models based on traditional demographic data by using only geolocated tweets. Our approach provides a method for discovering novel correlations between open-vocabulary topics and health variables, allowing researchers to discover yet unknown contributing factors based on large collections of data with minimal effort.

Our findings, when applying our method to AHD risk, diabetes prevalence and the risk of various types of cancers, using geolocated tweets from the US only, show that a large variety of health-related variables can be predicted with surprisingly high precision based solely on social media data. Furthermore, we show that our model identifies known and novel risk or protective factors in the form of topics. Both aspects are of interest to researchers and policy makers. Our model proved to be robust for the majority of targets it was applied to.

For AHD risk, we show that our approach significantly outperforms previous models based on topic models such as LDA or traditional statistical models \cite{eichstaedt2015psychological}, achieving a $\rho$-value of 0.46, an increase of 0.09 over previous approaches. For diabetes prevalence our model correctly predicts its geographic distribution by identifying linguistic features common in high-prevalence areas among other features, with a $\rho$-value of 0.73. For melanoma risk, it finds a high-correlation with the popularity of outdoor activities, corresponding to exposure to sunlight being one of the main risk factors in skin cancer, with an overall $\rho$-value of 0.72.

One of the main limitations of our approach is the need for a large collection of sentences for each community as well as a large number of communities with target variables, leading to potentially unreliable results when this is not the case, such as for social media posts by individuals or when modeling target values which are only available in e.g. few counties. Further research is needed to ascertain whether significant results can also be achieved in such scenarios, and if robustness of our approach is improved compared to bag-of-words-based baselines \cite{eichstaedt2015psychological,brown2018does,schwartz2018more}. Furthermore, all mentioned approaches rely on \emph{correlation}, and thus do not provide a way to determine any \emph{causation}, or ruling out of potential underlying factors not captured by the model. Even though using social media data  introduces a non-negligible bias towards users of social media, our approach was able to predict target variables tied to very different age-groups, which is encouraging and supports the robustness of our approach.
\\

Our method captures language features on a community scale. This raises the question of how these findings can be translated to the individual person. Theoretically, a community-based model as described above could be used to rank social media posts or messages of an individual user, with respect to specific health risks. However, as we currently do not have ground truth values on the individual level, and since user's social media history has very high variance, this is left for future investigation.

Future research should also address the applicability of our model to textual data other than Twitter and potentially from non-social media sources, to communities that are not geography based, to the time evolution of topics and health/lifestyle statistics, as well as to targets that are not health related. The general methodology offers promise for new avenues for data-driven discovery in fields such as medicine, sociology and psychology.

\paragraph{Acknowledgements.}
We would like to thank 
Ahmed Kulovic
and Maxime Delisle
for valuable input and discussions.

\clearpage

\bibliography{bibliography}
\bibliographystyle{acl_natbib}

\cleardoublepage
\appendix

\section{Appendices}

\subsection{Additional Figures}\label{sec:addCharts}

\begin{figure}[!htb]
\centering
\subfloat[]{
   \includegraphics[width=0.95\linewidth]{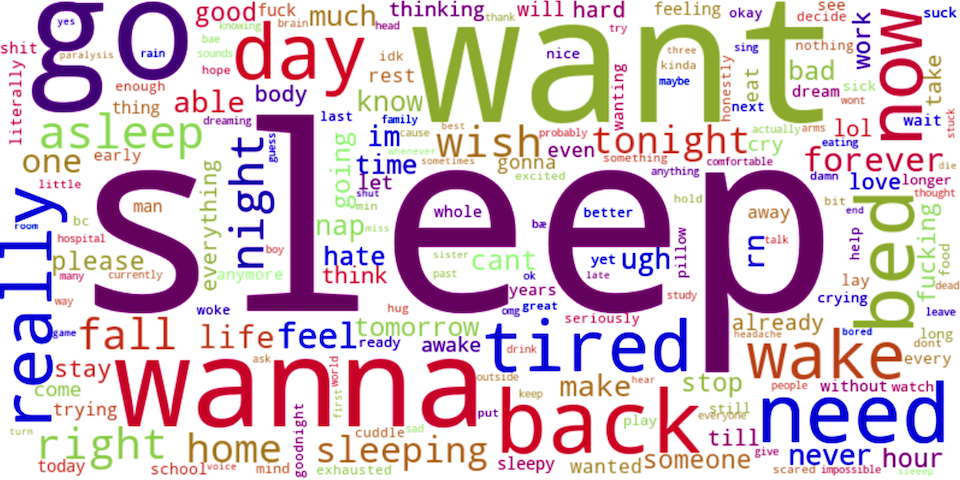}
}

\subfloat[]{
   \includegraphics[width=0.95\linewidth]{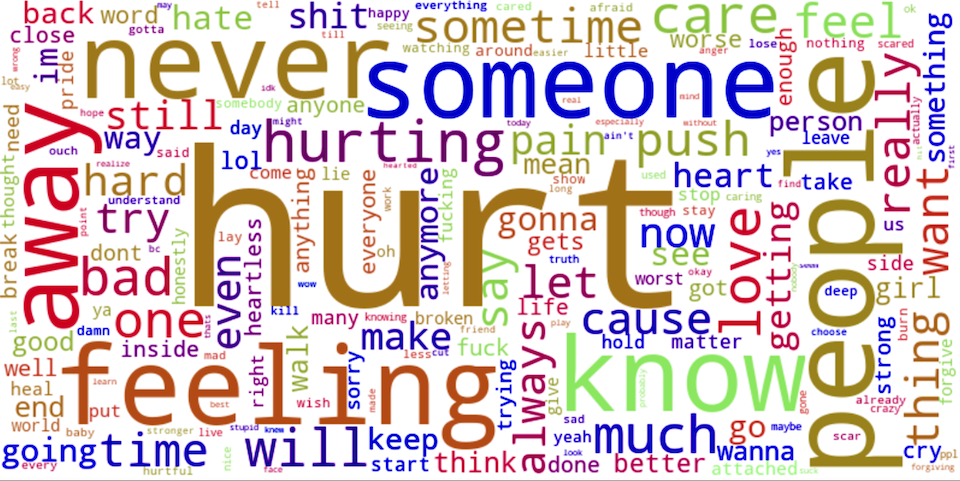}
}

\subfloat[]{
   \includegraphics[width=0.95\linewidth]{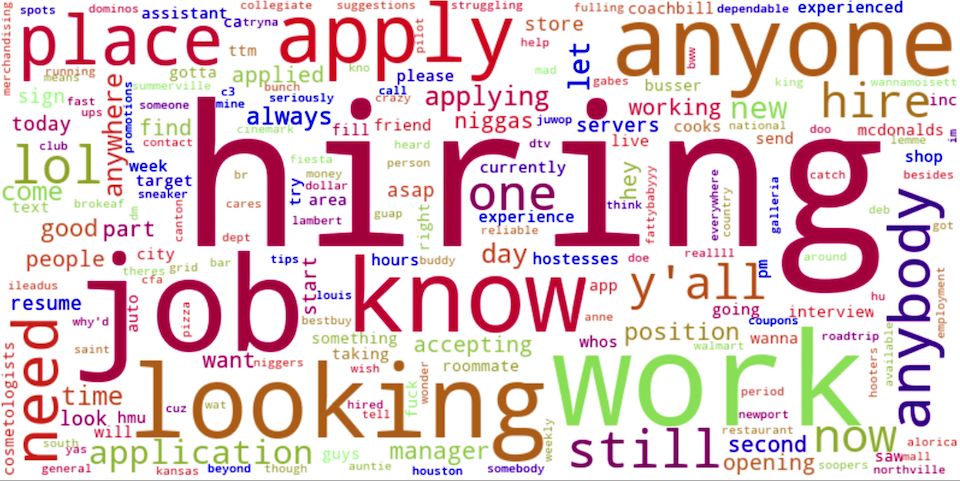}
}

\subfloat[]{
   \includegraphics[width=0.95\linewidth]{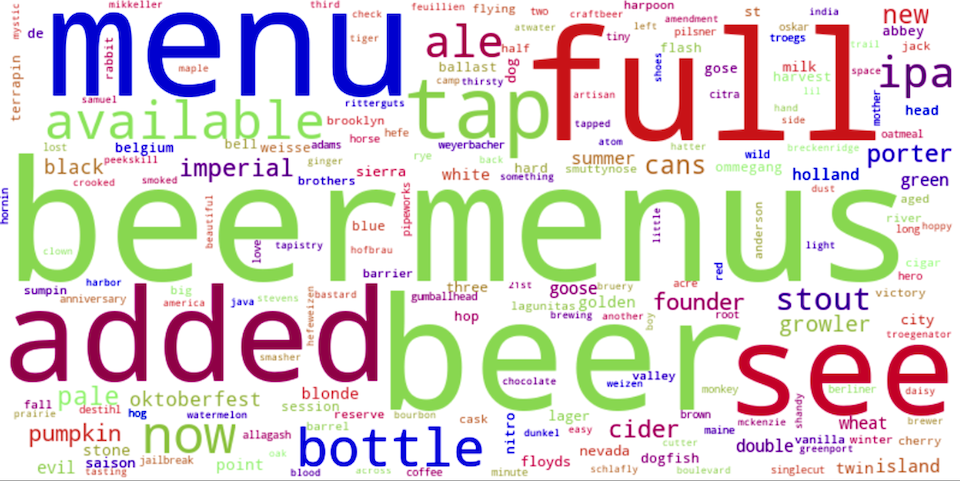}
}

\caption{Word clouds of topics correlating with \textbf{colorectal cancer}: (a) (b)strongest positively correlated topics (c) (d) strongest negatively correlated topics among $M=2000$ clusters.}
\end{figure}

\newpage

\begin{figure}[!htb]
   \centering
   \includegraphics[width=0.95\linewidth]{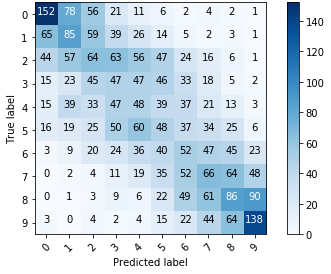}\vspace{-3mm}

    \caption{Confusion matrix for decile-based prediction of \textbf{diabetes} prevalence.}
\end{figure}
\vspace{3mm}

\subsection{Implementation Details}
Tweets were collected according to the provided datorium IDs using the Tweepy\footnote{https://www.tweepy.org/} library. The tweets were then imported into Google BigQuery\footnote{https://cloud.google.com/bigquery/} and processed using Apache Beam\footnote{https://beam.apache.org/}. The sentence embeddings were computed using the official Sent2Vec source code and the provided 700-dimensional pre-trained model for tweets (using bigrams)\footnote{https://github.com/epfml/sent2vec}. Clustering was performed by libKMCUDA\footnote{https://github.com/src-d/kmcuda}.
Scikit-learn\footnote{https://scikit-learn.org/stable/} was used for 10-fold cross validation, Ridge regression, calculating the correlation and hyperparameter search.

\end{document}